\documentclass[runningheads]{llncs}
\usepackage{graphicx}
\usepackage{comment}
\usepackage{multirow}

\title{NavTopo: Leveraging Topological Maps For Autonomous Navigation Of a Mobile Robot}
\author{Kirill Murayvev\inst{}\orcidID{0000-0001-5897-0702} \and
Konstantin Yakovlev\inst{}\orcidID{0000-0002-4377-321X}}
\authorrunning{K. Muravyev and K. Yakovlev}
\institute{Federal Research Center ``Computer Science and Control'' of Russian Academy of Sciences, Moscow, Russia
\\
\email{muraviev@isa.ru, yakovlev@isa.ru}
}

\begin{document}

\maketitle 

\begin{abstract}
    Autonomous navigation of a mobile robot is a challenging task which requires ability of mapping, localization, path planning and path following. Conventional mapping methods build a dense metric map like an occupancy grid, which is affected by odometry error accumulation and consumes a lot of memory and computations in large environments. Another approach to mapping is the usage of topological properties, e.g. adjacency of locations in the environment. Topological maps are less prone to odometry error accumulation and high resources consumption, and also enable fast path planning because of the graph sparsity. Based on this idea, we proposed NavTopo – a full navigation pipeline based on topological map and two-level path planning. The pipeline localizes in the graph by matching neural network descriptors and 2D projections of the input point clouds, which significantly reduces memory consumption compared to metric and topological point cloud-based approaches. We test our approach in a large indoor photo-relaistic simulated environment and compare it to a metric map-based approach based on popular metric mapping method RTAB-MAP. The experimental results show that our topological approach significantly outperforms the metric one in terms of performance, keeping proper navigational efficiency.
\end{abstract}

\section{Introduction}

%$$\mathbf{x} = [p, \theta, v, w]^T$$

\begin{figure}[ht]
    \centering
    \includegraphics[width=0.9\textwidth]{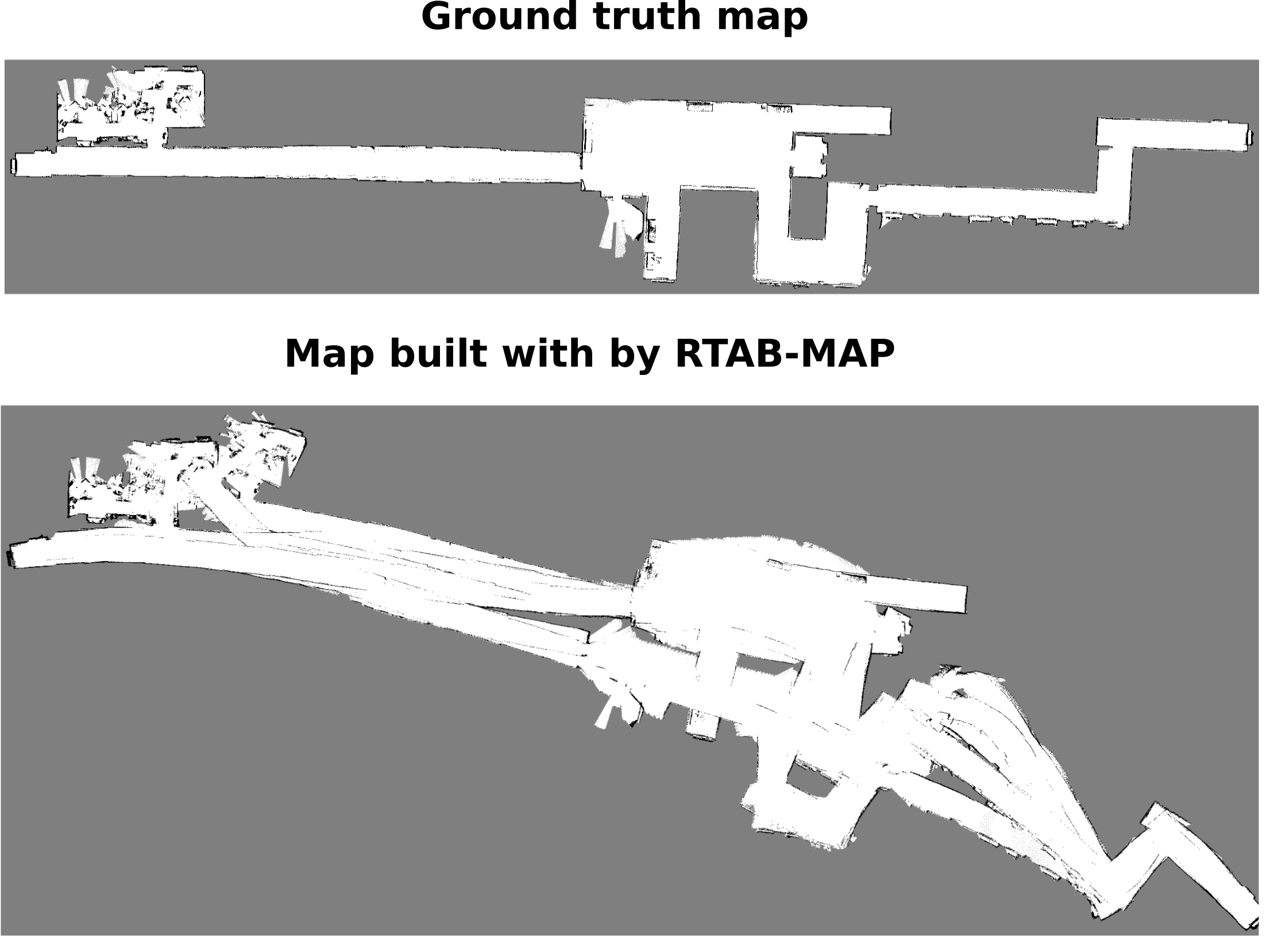}
    \label{fig:corridor_bifurcation}
    \caption{An occupancy grid built by RTAB-MAP \cite{labbe2019rtab} algorithm for a large corridor with noised odometry, compared to the ground truth occupancy grid. In the middle and in the left end of the map, a corridor is mapped twice.}
\end{figure}

Autonomous navigation is a crucial property for a mobile robot operation. In many cases, mobile robots solve tasks related to navigation in unknown or chaniging environment. It is a challenging task which is divided to map building, localization, path planning, and path following. A map of the environment is usually built as a 2D occupancy grid or a 3D voxel grid. Such metric environment representation is convenient for computation and contains complete information about obstacles. However, in large environments, consumption of memory and computational resources may be too high for online map maintaining. Also, planning path in large grids may take long time. One more drawback of metric maps is the accumulation of odometry error, which may lead to incorrect mapping, like "corridor bifurcation" shown in Fig. \ref{fig:corridor_bifurcation}.

Another approach to environment representation is the usage of topological structures like graph of locations. In such topological structures, vertices represent some places in the environment, and edges connect neighbour places. Mapping only topological properties of the environment significantly reduces memory consumption of the map and also mitigates odometry error accumulation \cite{schmid2021unified}. Also, path planning in sparse topological maps is much faster than in dense metric maps \cite{gomez2020hybrid}.

In this work, we introduce NavTopo -- a full pipeline of mobile robots navigation based on topological maps. The mapping and localization parts of the pipeline are based on the PRISM-TopoMap \cite{muravyev2024prism} method. This method builds a graph of locations by raw perception and odometry measurements. The locations in the graph represent areas captured by a scan from a point traveled by the robot, and edges connect locations in case of adjacency (i.e. two locations overlap by a drivable area). The proposed navigation system uses two-level path planning. First, the high-level topological path is searched between the location the robot is located in and the location related to the goal. Second, a local metric path is planned to the center of the next location on a local metric occupancy grid. This local grid is created from a union of the current location and its neighbours in the graph.

To evaluate the proposed navigational system, we carried out experiments in a large photorealistic simulated environment. During the experiments, we compare the NavTopo pipeline with the metric approach to navigation based of RTAB-MAP metric mapping method. Both approaches are evaluated using performance metrics such as memory consumption and path planning time, and navigational efficiency. The experiments show that the proposed approach significantly outperforms metric map-based approach in terms of performance and just slightly inferior in navigational efficiency because of the graph sparsity.

\section{Related Work}

\subsection{Metric Mapping}

The problem of metric map building and Simultaneous Localization and Mapping (SLAM) has long history of study. Pioneering SLAM methods track robot's position and surrounding objects using Extended Kalman Filter (EKF) \cite{cheeseman1987stochastic} to estimate robot's trajectory from noised sensor data. More recent works extract features from input images and track robot motion matching these features \cite{mur2015orb}, or match input images directly using photometric error \cite{engel2014lsd}. Methods like RTAB-MAP \cite{labbe2019rtab} or Cartographer \cite{hess2016real} build dense 2D and 3D maps from input point clouds using probabilistic optimization and loop closure techniques. Such methods are widely used because of the grids built by them are convenient for path planning. However, the computations and memory consumption of these methods is significant \cite{muravyev2022evaluation}.

In last years, the progress in deep learning techniques lead to creation of neural network-based SLAM methods like \cite{teed2021droid}, \cite{min2021voldor}. In method \cite{min2021voldor} neural network is used for optical flow estimation in order to accurately estimate robot's position. In work \cite{teed2021droid} SLAM is performed by a fully learning-based pipeline, including feature extraction and bundle adjustment. Both these methods build dense map and have small error, however, they require powerful GPU to operate, so use of these methods in a mobile robotic system is difficult.

\subsection{Topological Mapping}

The topological mapping is an actively developing research field, and it encompasses various methods which differ in graph type and input data. For example, some methods like \cite{blochliger2018topomap}, \cite{chen2022fast} build a topological map offline for fast path planning using a pre-built metric map. Some other methods like \cite{hughes2022hydra}, \cite{bavle2022s}, build hybrid multi-level metric-topological map. Such hybrid map provides complete and detailed information about the environment, however, these mapping methods are susceptible to metric SLAM's shortcomings like high resources consumption and odometry error accumulation.

In recent years, a large number of learning-based topological mapping methods which do not rely on a metric map was emerged. However, most of them (like \cite{kim2023topological}, \cite{kwon2021visual}) are designed for solving specific task like image-goal navigation in simulated environments and were not tested on data from a real robot. Also, such methods usually rely on neural network-predicted features only, which can cause linking far locations in the environment and result in navigation failures. Some other learning-based methods like \cite{wiyatno2022lifelong} work in real environments with edge filtering, but require a pre-built topological map at start.

In our previous work \cite{muravyev2024prism}, we proposed a PRISM-TopoMap method which builds a graph of locations online without a global metric map and without relying on global metric coordinates. Because of use of only local odometry and correction by scan matching, our method does not suffer from odometry error accumulation. Also it provides very fast path planning because of the graph sparsity. We choose this method as a base of our navigation pipeline.

\subsection{Path Planning}

One of the most common approaches to robot trajectory planning is to represent the environment as a graph and find paths in this graph. In case of a metric map, the vertices of the graph usually form a regular grid \cite{elfes1989using} of free space cells. In case of a topological map (e.g. graph of locations), the path is planned in it directly. In topological graphs, a common graph path finding algorithms like Dijkstra \cite{dijkstra1959note} or A* \cite{hart1968formal} are widely used. For grid-like graphs, besides these common algorithms, any-angle path planning methods like Theta* \cite{daniel2010theta} or D*-lite \cite{koenig2005fast} are used. Also, in partially unknown environments, probabilistic approaches like RRT \cite{lavalle2001rapidly} or RRT* \cite{karaman2011sampling} are widely used to plan paths in uncertainty conditions.

Many topological navigation approaches suppose two-level path planning, with a global path planned in the topological graph, and a local path planned in order to reach the next vertex on the global path. For example, in method \cite{schmid2021unified} global path planning by A* algorithm is used to reach frontiers of the map, and local path planning by an RRT-based approach is used to search fine path in a sliding window near the robot. In the method \cite{gomez2020hybrid} a similar scheme is used -- first, global path in the graph of rooms is planned to the room node where the goal is located. Next, a local path between nodes is searched using RRT algorithm.

In recent years, end-to-end learning-based approaches expanded the field of study related to robots navigation. For example, RL-based navigation approaches like DD-PPO \cite{ddppo} are widely used for point-goal navigation, but mostly in simulated environments. In the methods \cite{kim2023topological} and \cite{kwon2021visual} image-goal navigation task is solving by and end-to-end RL-based approach with building of the graph of locations. In these methods, two-level approach is used: high level RL policy chooses a location where the goal located in, and low-level policy outputs actions which are needed to reach the target location. Some methods like SkillFusion \cite{staroverov2023skill} involve combined approach to navigation and switch between classical and learning-based methods depending from conditions of the map.

\section{System Overview}

\begin{figure}[ht]
    \centering
    \includegraphics[width=1.0\textwidth]{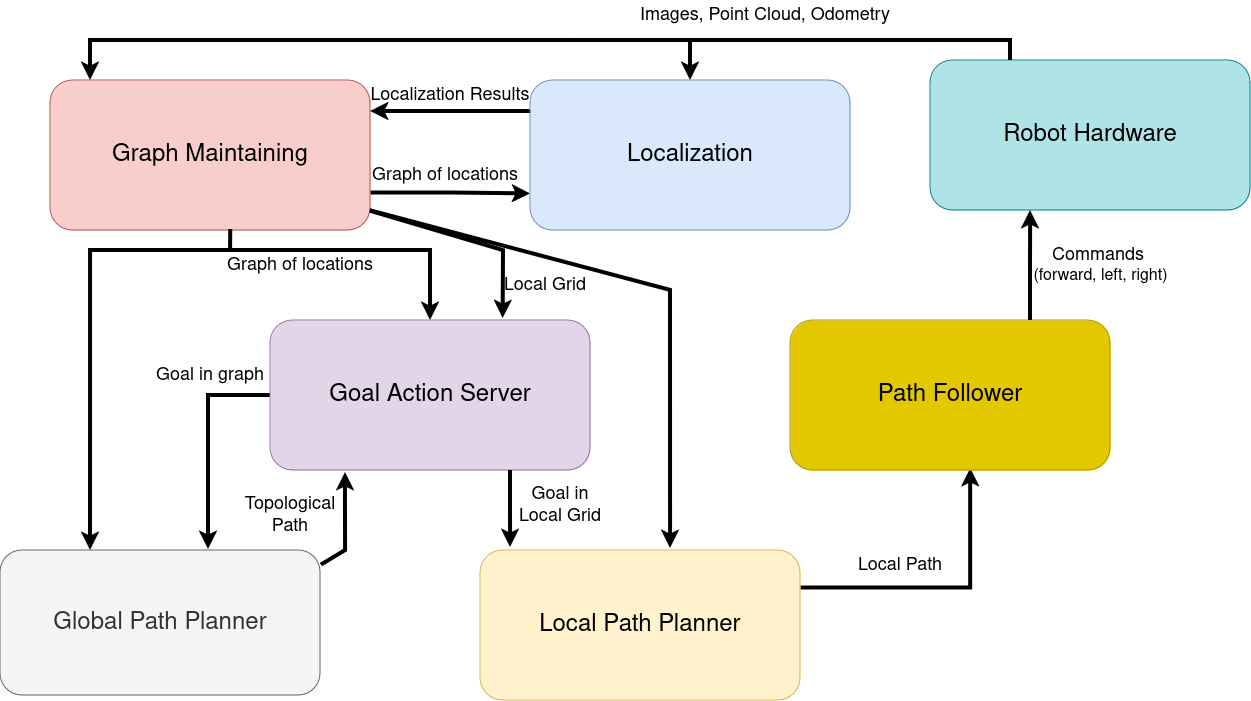}
    \label{fig:method_scheme}
    \caption{A scheme of the proposed navigation pipeline}
\end{figure}

The proposed system maintains a graph of locations of the environment for efficient robot navigation. Its scheme is shown in Fig. \ref{fig:method_scheme}. It is divided into several modules: mapping and localization, global and local path planners, and local path follower. Graph maintaining module maintains and outputs the graph of locations, and localization module helps to find the current state in the graph. Global path planner finds path by locations in the graph, whereas local path planner finds local metric path in order to reach the next location. Path follower generates robot motion commands in order to move robot along the local path. A goal action server module handles graph of locations and local occupancy grid and distributes tasks to planners in order to create paths from the current robot's position to the goal.

\subsection{Graph Structure}
\begin{figure}[ht]
    \centering
    \includegraphics[width=0.6\textwidth]{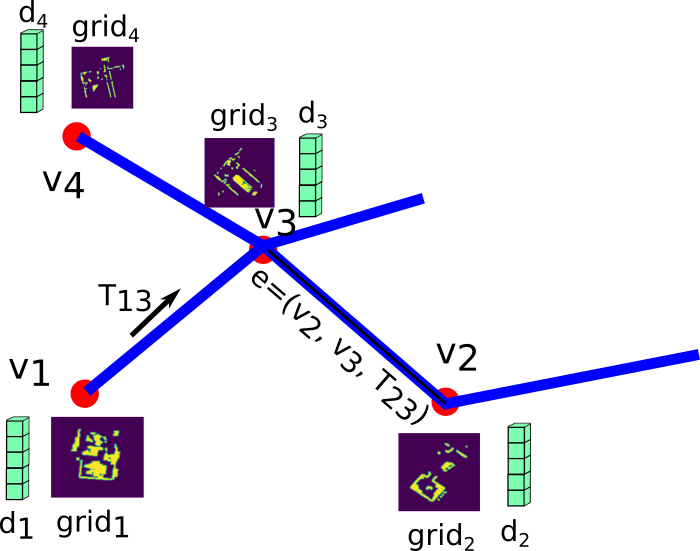}
    \label{fig:graph_structure}
    \caption{A graph of locations built by the NavTopo method}
\end{figure}

The proposed system builds and maintains a graph of locations $G=(V,E)$ for navigation. Each location in the graph $v \in V$ is assigned its occupancy grid for local path planning, and a descriptor for neural network-based localization: $v = (grid_v, desc_v)$. Two locations are connected by an edge if they are adjacent (i.e. free areas of their occupancy grids are connected). Each edge is assigned the locations it connects and a relative pose between the observation points of the locations it connects: $e = (u, v, T_{uv})$. An example of the graph is shown in Fig. \ref{fig:graph_structure}.

\subsection{Graph Maintaining}
\begin{figure}[ht]
    \centering
    \includegraphics[width=0.65\textwidth]{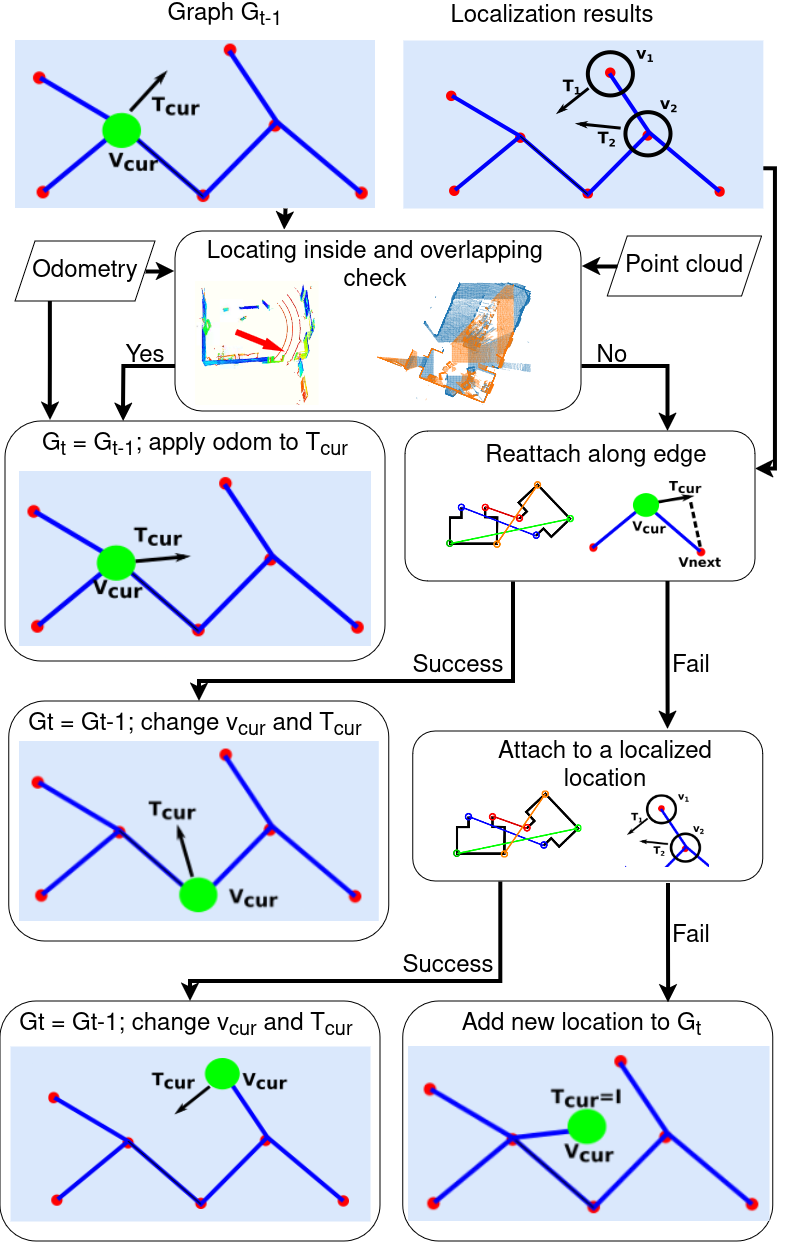}
    \label{fig:graph_maintaining_scheme}
    \caption{A scheme of the graph maintaining module}
\end{figure}

The graph maintaining module is based on the graph maintaining module of the PRISM-TopoMap method \cite{muravyev2024prism}. It builds and expands graph of locations and also maintains the current robot's state in the graph using observations from robot's perception and odometry sensors and output of the localization module. At each step $t$, the module outputs the localtion $v_{cur}^t$ where the robot is currently located and the relative robot's pose $T_{cur}^t$ in this location. The values of $v_{cur}^t$ and $T_{cur}^t$ are updating depending from the odometry changes, localization, and route to the goal. The scheme of the graph maintaining module is shown in Fig. \ref{fig:graph_maintaining_scheme}.

The process of graph update consists of the following stages. First, we check whether the robot is still inside $v_{cur}^{t-1}$. If it is inside, we just change $T_{cur}^t$ from the odometry observations. Otherwise, we try to change $v_{cur}$'s value to one of the neighbors of $v_{cur}^{t-1}$ (i.e. move along an edge). If the robot is tasked to move to a goal, we try to change $v_{cur}$ to the next node in the path to the goal. If the change is unsuccessful, we remove the edge we tried to change along, and try to change $v_{cur}$ to one of the nodes from the localization module output. If we still failed to change $v_{cur}$, we add new location into the graph and set $v_{cur}^t$'s value to it.

The process of checking being of the robot in a location and computation of $v_{cur}^t$ and $T_{cur}^t$ is detailly described in the work \cite{muravyev2024prism}. To check being inside a location, we match (find a relative pose) the current robot's scan and the scan of the location using a 2D feature-based scan matching technique and also calculate overlapping of the scans. If the overlap is low, the robot is also considered outside the location.

\subsection{Localization}

\begin{figure}[ht]
    \centering
    \includegraphics[width=1.0\textwidth]{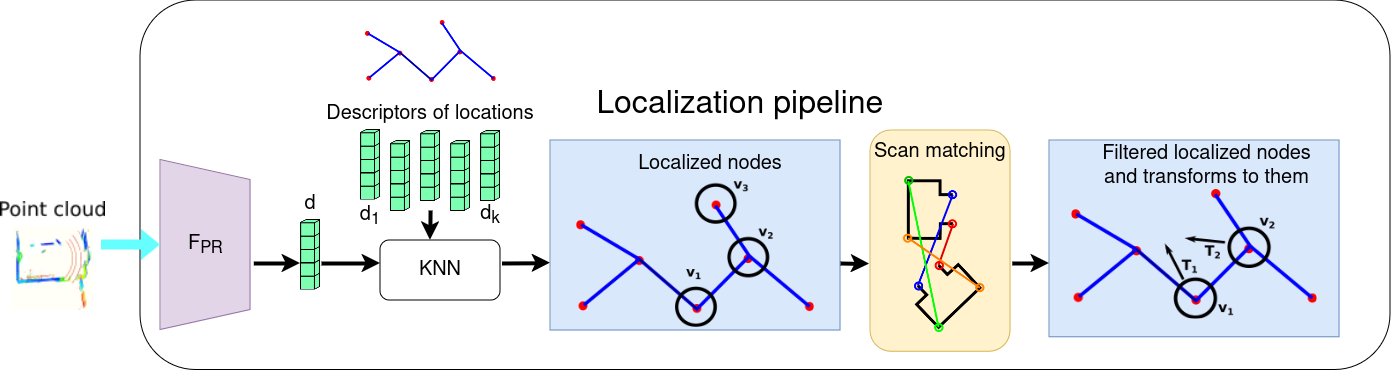}
    \label{fig:localization_scheme}
    \caption{A scheme of the localization module}
\end{figure}

For localization, we use two-stage pipeline: first, we find possible robot's positions in the graph via place recognition technique, and next we filter the found locations and find relative poses from robot to them using a scan matching technieque. The scheme of the localization module is shown in Fig. \ref{fig:localization_scheme}. For place recognition, we use MinkLoc3D model \cite{komorowski2021MinkLoc3DPointCloud}, which is based on raw point clouds. The model predicts descriptor from a point cloud, and similarity between point clouds is measured as proximity of their descriptors. We store these descriptors for all the location in the graph. To localize the robot in the graph, we choose five descriptors of the graph's locations which are closest to the descriptor of the current robot's scan.

To eliminate localization errors and calculate relative poses between locations, we match the found point cloud scans based on features extracted from their 2D projections. So, we can store only descriptors and 2D projections in the graph, without raw point clouds, that significantly reduces memory and computational resources consumption.

\subsection{Path Planning}
\begin{figure}
    \centering
    \includegraphics[width=0.8\textwidth]{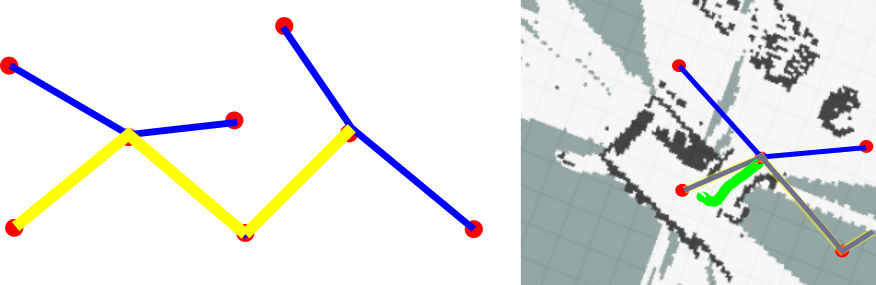}
    \caption{Two-level path planning used in NavTopo approach: high-level global topological path (left, yellow) and low-level local metric path (right, green)}
    \label{fig:path_planning}
\end{figure}

To plan paths to a goal, we use a two-level approach. At a high level, we plan global path in the graph of locations -- from $v_{cur}$ to the location where the goal is located in. This path is planned using Dijkstra's algorithm \cite{dijkstra1959note}, with edges weighted by lengths of these relative poses. At a low level, we plan the local path to reach the next point on the global path and avoid collisions with obstacles. A scheme of the path planning module is shown in Fig. \ref{fig:path_planning}. If the goal is located in $v_{cur}$, the next point is the goal position relative to $v_{cur}$'s observation point. Otherwise, it is the position from a relative pose written on the first edge of the path.

To dispatch global and local path planners for navigation to the goal, we use the goal action server. This server receives a goal in the format $(v_{goal}, T_{goal})$, where $v_{goal}$ is the index of location the goal located in, and $T_{goal}$ is the goal's position relative to the observation point of $v_{goal}$. Also, during the navigation process, the server receives the current state in the graph $(v_{cur}, T_{cur})$ and global path $P=(e_1, \dots, e_n); e_1 = (v_{cur}, v_1, T_1), e_2 = (v_1, v_2, T_2), \dots, e_n = (v_{n-1}, v_n=v_{goal}, T_n)$. Using this data, the server checks proximity between robot and goal: if $v_{cur}=v_{goal}$ and $|T_{cur} - T_{goal}| < \varepsilon$, then the goal is considered reached, otherwise the server gives task to the local path planner to reach the next point $T_{next}$ on the global path. If $v_{cur}=v_{goal}$ then $T_{next} = T_{goal}$. Otherwise, $T_{next}=T_1$.

\subsection{Path Following}

To move the robot to the goal, we follow the local path using a simple and straightforward algorithm described in the work \cite{muravyev2021enhancing}. The algorithm checks the difference between the robot's orientation angle and the direction to the next point of local path. If the angle is low, the robot is moved forward. Otherwise, the robot is turned left or right to eliminate this angle difference.

\section{Experiments}

To prove the navigational and computational efficiency of our navigation system, we performed some experiments in a large indoor photorealistic simulated scene. On this scene, we compared the proposed system with the metric SLAM-based approach. For both metric and our topological solutions, we measured memory consumption, path planning time and navigational efficiency.

\subsection{Setup}

\begin{figure}
    \centering
    \includegraphics[width=1.0\textwidth]{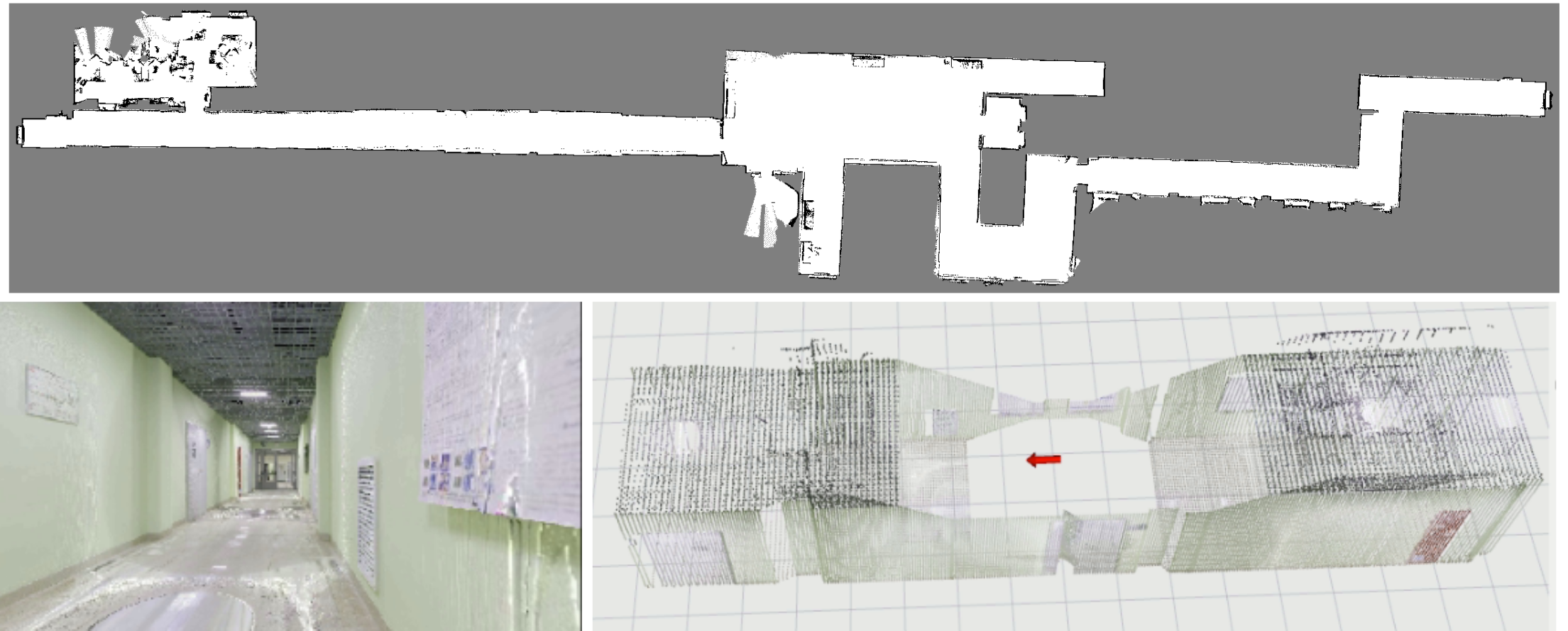}
    \caption{The ground-truth 2D scene map and the scene view: image and point cloud}
    \label{fig:dataset_sample}
\end{figure}

We tested our approach and the metric approach on in Habitat simulator \cite{savva2019habitat}. The scene for evaluation is created by ours from a long branched corridor with total length of 150 m.  The map of the scene and observations are shown in Fig. \ref{fig:dataset_sample}.

Both metric and topological navigational pipelines were fed with point clouds and precise odometry data from the simulator. First, a virtual agent moved along the environment by a pre-defined 1 km route and both algorithms were tasked to build a map on the fly. Next, both algorithms loaded the map built on the first stage, and were tasked to move a virtual agent between different start and goal positions. Totally, 20 runs from start to goal were performed, and the average distance between a start and a goal was 78 m. During the navigation, memory consumption was measured, as well as the path planning time and the navigational efficiency. The navigational efficiency was measured as a ratio between the shortest path from start to goal and the traveled path.

\subsection{Metric SLAM}

As a metric SLAM method for comparison we chose RTAB-MAP \cite{labbe2019rtab} as one of the most popular and convenient solutions for navigation using occupancy grid. At a stage of map building, RTAB-MAP builds global occupancy grid map for the whole environment and saves it to the data base. The occupancy grid resolution was chosen to 0.1 m. As an input of RTAB-MAP, only point clouds and odometry data are used. At a stage of navigation, path to a goal position is planned on RTAB-MAP's grid using Theta* path planner. For path following, the same algorithm is used as in our topological approach.

\subsection{Results}

\begin{figure}
    \centering
    \includegraphics[width=1.0\textwidth]{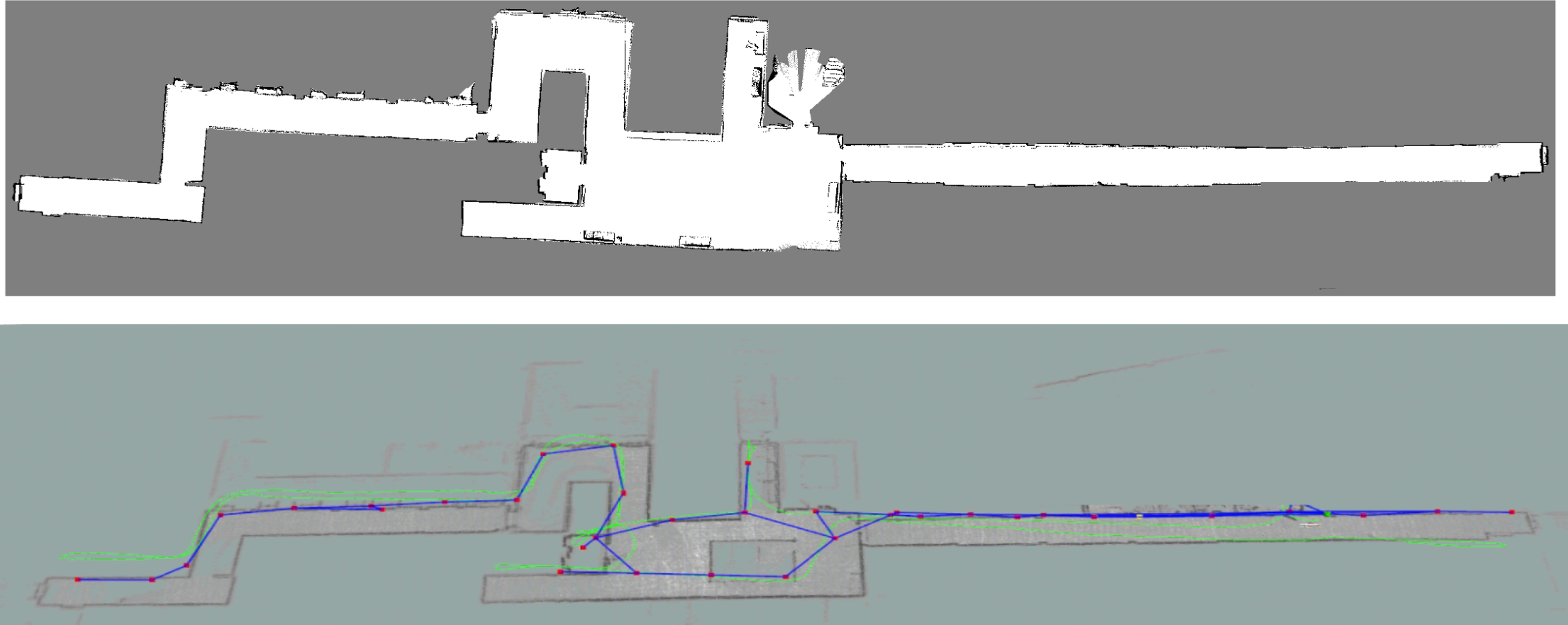}
    \caption{Map built by RTAB-MAP (top) and map built by our NavTopo method (bottom), aligned with ground truth occupancy grid map and ground truth trajectory}
    \label{fig:results}
\end{figure}

The results of the experiments are shown in the Table \ref{tab:results}. The maps built by RTAB-MAP and by the NavTopo method are shown in Fig. \ref{fig:results}.

\begin{table}[h]
    \centering
    \begin{tabular}{|l|c|c|c|}
         \hline
          & Map memory consumption (MB) & Planning time (ms) & Efficiency \\
         \hline
        NavTopo & 57 & 6 & 0.89 \\
        RTAB-MAP & 352 & 830 & 0.95 \\
        \hline
    \end{tabular}
    \caption{Performance and efficiency values of our NavTopo method and RTAB-MAP method}
    \label{tab:results}
\end{table}

As seen in the table, the proposed NavTopo method consumed 6 times less memory than RTAB-MAP. Also, the path planning times with NavTopo were more than two orders less than the planning times with a global RTAB-MAP's occupancy grid. However, because of sparsity of a graph of locations, navigational efficiency reached only 0.89, compared to 0.95 with global metric maps.  Smoothing paths in graphs of locations is planned in our future research.

\section{Conclusion and Future Work}

In this work we proposed NavTopo -- a system for autonomous robot navigation based on a graph of locations and two-level path planning. The proposed system maintains the graph using 2D projections of point clouds and feature descriptors predicted from them. So, raw point clouds or 3D maps are not stored the graph, which significantly reduces memory consumption. The experiments conducted in a large indoor photorealistic environment showed that the proposed navigation pipeline significantly outperforms metric approach in terms of performance and path planning speed, with insignificant drop in navigational efficiency. Moving forward, we plan to carry out extensive experiments with noised odometry data, and test the proposed pipeline on a real robot. Also, we plan to develop novel classical and RL-based path followers for smooth trajectory generation and obstacle avoidance.

\subsubsection{Acknowledgements}
The reported study is supported by the Ministry of Science and Higher Education of the Russian Federation under Project 075-15-2024-544.

\bibliographystyle{splncs03_unsrt}
\bibliography{mybibliography}

\end{document}